\documentclass[5p,times,twocolumn,nopreprintline]{elsarticle}
\usepackage[T1]{fontenc}
\usepackage[utf8]{inputenc}
\usepackage{amsmath,amssymb,booktabs}
\usepackage{xcolor}
\usepackage{graphicx}
\usepackage{algorithm}
\usepackage{algpseudocode}
\usepackage{tabularx}
\usepackage{tikz}
\usetikzlibrary{arrows.meta}
\usepackage[bookmarks=false]{hyperref}
\hypersetup{colorlinks=true,linkcolor=blue,citecolor=blue,urlcolor=blue}
\biboptions{numbers,sort&compress}

\usepackage{etoolbox}

\makeatletter
\newcommand{\StateWrap}[1]{%
  \State\parbox[t]{\dimexpr\linewidth-\ALG@thistlm\relax}{%
    \raggedright\strut #1\strut
  }%
}
\makeatother

\algrenewcommand{\algorithmicrequire}{\textbf{Input:}}
\algrenewcommand{\algorithmicensure}{\textbf{Output:}}

\newcommand{\ROSEDisplaySpacing}{%
  \setlength{\abovedisplayskip}{6pt plus 1pt minus 1pt}%
  \setlength{\belowdisplayskip}{6pt plus 1pt minus 1pt}%
  \setlength{\abovedisplayshortskip}{3pt plus 1pt minus 1pt}%
  \setlength{\belowdisplayshortskip}{5pt plus 1pt minus 1pt}%
}

\AtBeginDocument{%
  \apptocmd{\normalsize}{\ROSEDisplaySpacing}{}{%
    \PackageWarning{ROSE}{Could not patch normal-size equation spacing}%
  }%
  \normalsize
  \setlength{\jot}{2pt}%
}

\renewenvironment{frontmatter}{}{\maketitle}

\begin{document}
\begin{frontmatter}
\title{Sample-efficient prediction of high-entropy alloy strength through outcome-aware robust sequential learning}
\author[inst1]{Ahmed Shoyeb Raihan}
\author[inst2]{Md Habibor Rahman}
\author[inst1]{Ashish Nimbarte}
\author[inst1]{Imtiaz Ahmed\corref{cor1}}

\cortext[cor1]{Corresponding author.}
\ead{imtiaz.ahmed@mail.wvu.edu}

\address[inst1]{Department of Industrial and Management Systems Engineering,
West Virginia University, Morgantown, 26506, West Virginia, USA}

\address[inst2]{Department of Mechanical Engineering,
University of South Carolina, Columbia, 29208, South Carolina, USA}
\begin{abstract}
Accurate prediction of high-entropy alloy (HEA) strength can guide composition and processing decisions; however, obtaining tensile-test data for HEAs can be costly and experimentally demanding. This study proposes Robust Outcome-aware Sequential Experimentation for High-Entropy Alloys (ROSE-HEA), a sample-efficient framework for predicting ultimate tensile strength (UTS) from alloy composition and processing conditions. Its acquisition policy anticipates how possible UTS outcomes update predictions and the support for complementary Gaussian process (GP) descriptions. It then reassesses the worst-case expected predictive risk remaining across a reference domain to select informative candidates. A retrospective evaluation on 501 HEA records compares ROSE-HEA with expected-improvement (EI) and upper-confidence-bound (UCB) Bayesian optimization (BO), random GP sampling, and eight conventional regressors across 30 matched runs with shared initial observations and test conditions. Starting from 25 initial observations and acquiring 100 additional responses, ROSE-HEA achieves the lowest mean root-mean-square error (RMSE), at 152.84~MPa, and the highest mean $R^2$, at 0.7443. Mean RMSE is reduced by 6.16\% and 6.37\% relative to EI and UCB, respectively. The selected observations improve prediction across six regressor families and provide broader design-space coverage. ROSE-HEA also surpasses the strongest conventional regressor's mean predictive performance using 125 rather than 185 observations, corresponding to 32.4\% fewer training observations. These findings demonstrate the potential of outcome-aware robust sequential learning to improve UTS prediction in HEAs under constrained training budgets.
\end{abstract}
\end{frontmatter}

\section{Introduction}
\label{sec:introduction}

High-entropy alloys (HEAs) have opened a broad design space for structural materials by combining multiple principal elements in substantial concentrations~\cite{Yeh2004HEA,Cantor2004}. Their chemical flexibility creates opportunities to tailor microstructures and mechanical performance through coordinated alloying and thermomechanical processing~\cite{MiracleSenkov2017,George2019HEA}. Demonstrations of high damage tolerance, favorable strength--ductility combinations, and precipitation strengthening in selected HEAs highlight their potential for structural applications~\cite{Gludovatz2014Cryogenic,Li2016Metastable,He2016Precipitation}. Translating these advances into manufacturing decisions requires predictive relationships that connect controllable composition and processing choices to the properties of the resulting material.

Predicting the tensile strength attainable from a given composition and processing route is important for evaluating HEAs for structural applications. Ultimate tensile strength (UTS), defined as the maximum engineering stress sustained during tensile testing, provides a common basis for comparing alloy strength. Alloy chemistry and thermomechanical history jointly influence phase constitution, grain structure, and deformation mechanisms, producing substantial variation in UTS~\cite{George2020Mechanical,Otto2013Tensile}. Homogenization, cold working, and annealing can consequently change the strength attained by a given composition, while their effects vary across alloy systems. Learning these relationships enables UTS prediction at untested conditions, helping assess processing alternatives and prioritize mechanical characterization. Accurate predictive models can also support new alloy development by identifying promising compositions and processing combinations for experimental validation~\cite{Wang2023HEAStrength}.

Data-driven modeling offers a flexible route to constructing this predictive relationship. Machine learning (ML) can extract nonlinear trends and interactions from measured material responses, complementing physical understanding of complex multicomponent systems~\cite{Schmidt2019Review,Rickman2019Informatics}. HEA studies have demonstrated the value of ML-assisted property design~\cite{Wen2019HEAML}, while composition- and processing-aware neural modeling has enabled tensile-property prediction and experimentally validated alloy development~\cite{Wang2023HEAStrength}. These advances support a transition from repeated trial-and-error investigations toward model-informed material and process selection. Achieving reliable predictions with a practical experimental budget remains a central requirement for extending their manufacturing value.

For HEAs, acquiring a new UTS measurement can involve alloy preparation, thermomechanical treatment, specimen fabrication, and tensile testing. Exploring the many combinations of elemental fractions and processing conditions therefore places substantial demands on experimental resources. Under a restricted budget, predictive performance depends on both the model and the information provided by the measured conditions \cite{Raihan2026CASMART}. Randomly acquired observations can repeatedly characterize similar conditions while leaving other alloy compositions and processing conditions poorly represented. Greater sample efficiency requires acquiring observations that resolve the model's current knowledge gaps and improve subsequent
predictions.

Sequential learning (SL) addresses this requirement by linking predictive modeling to adaptive candidate selection. Starting from a small initial set, an SL framework fits a model, identifies a useful candidate, acquires its response, and updates the model before making the next decision~\cite{Cohn1996AL,Lookman2019ActiveLearning}. Each measured response can thereby influence the allocation of the remaining budget. Adaptive materials-design studies have demonstrated the value of this feedback between predictions, uncertainty, and experimental observations~\cite{Balachandran2016Adaptive,Xue2016AdaptiveDesign}. Related manufacturing applications have also shown how sequential selection can improve learning of process responses and defect behavior with limited labeled observations~\cite{Raihan2024Surprise,Raihan2025Defect}. Bayesian optimization (BO) provides an established architecture for organizing such sequential decisions through two interacting components: a probabilistic surrogate model and an acquisition function~\cite{Shahriari2016HumanLoop}. The surrogate approximates the response from the available observations and quantifies predictive uncertainty; the acquisition function uses this information to rank the remaining experimental candidates. Gaussian processes (GPs) are widely used surrogates because they combine flexible nonlinear prediction with analytical uncertainty updates~\cite{RasmussenWilliams2006GPML}. Expected improvement (EI) and upper confidence bound (UCB) are established acquisition rules that use predicted responses and uncertainty to guide selection~\cite{Jones1998EGO,Srinivas2010GPUCB}. Their effectiveness for HEA strength prediction depends on how well successive selections improve predictions across relevant alloy compositions and processing conditions.

An important challenge in sample-efficient HEA modeling is determining which experiment will most improve UTS prediction across the relevant composition--processing space. The information provided by an observation extends beyond its input condition: through relationships learned by the surrogate, it can also refine predictions for other compositions and processing conditions. Its value therefore depends on both the uncertainty it resolves and the range of conditions that benefit from the observation. This assessment becomes more challenging when limited observations support several plausible predictive models. A new observation can change both their predictions and the relative confidence assigned to them~\cite{Riis2022FullyBayesianGP}. Prediction-oriented active learning provides a foundation for evaluating these effects through their influence on subsequent predictions~\cite{BickfordSmith2023EPIG}. For HEAs, the available observations may also represent some composition--processing regions more densely than others, leaving substantial uncertainty in sparsely characterized conditions. Protecting predictive quality across this space requires considering how much importance is assigned to different conditions. Distributionally robust selection addresses this issue by allowing these importance weights to vary within a prescribed range and evaluating the largest resulting predictive risk~\cite{Takeno2025DRAL}. As observations reduce uncertainty in one region, another can become the main source of remaining uncertainty. An effective acquisition policy should therefore anticipate both the information gained from an observation and the uncertainty pattern expected to remain afterward.

To address this challenge, this study proposes \emph{Robust Outcome-aware Sequential Experimentation for High-Entropy Alloys} (ROSE-HEA), a novel sequential learning framework centered on an \emph{outcome-aware robust risk-reduction} acquisition mechanism. The central idea is to select a candidate according to how effectively its anticipated information reduces the remaining predictive risk across relevant alloy conditions. Here, robustness addresses uncertainty in the relative importance of these conditions by assessing predictive risk under the most unfavorable weighting within a prescribed range. For each candidate, ROSE-HEA considers possible UTS outcomes and estimates how they would update strength predictions and the relative support for complementary GP models. These outcome-dependent updates quantify the expected learning benefit throughout a reference composition--processing space. The acquisition then reassesses which reference conditions would dominate the residual risk after this expected benefit is incorporated. Consequently, the emphasis assigned to different conditions adapts to the anticipated effect of each candidate experiment. By linking outcome-dependent predictive updates with this residual-risk reassessment, the policy directs selection toward observations expected to remain informative as the uncertainty pattern evolves. Within the ROSE-HEA framework, repeated acquisition and model updating progressively improve strength prediction using a limited number of training observations. The main contributions of this study are summarized as follows:
\begin{enumerate}
    \item We propose a novel outcome-aware robust acquisition mechanism that evaluates expected reference-wide learning gains and reoptimizes the weights governing residual predictive risk.

    \item We develop ROSE-HEA for sample-efficient HEA strength prediction and demonstrate lower mean prediction errors than EI-BO, UCB-BO, Random GP, and eight conventional regressors across 30 matched runs.

    \item We demonstrate the transferable predictive value of ROSE-HEA-selected observations through improved performance across six conventional regressor families.

    \item We analyze design-space coverage and regional accuracy, connecting the selected observations with broader input-space representation and improved prediction at initially distant conditions.

    \item We quantify sensitivity to initial and sequential training budgets and demonstrate improved prediction with fewer observations than conventional regressors.
\end{enumerate}

The remainder of the paper is organized as follows. Section~\ref{sec:related_work} reviews related work and research gaps. Section~\ref{sec:methodology} presents the ROSE-HEA framework, and Section~\ref{sec:dataset_experimental_protocol} describes the dataset and experimental protocol. Section~\ref{sec:results_discussion} presents and discusses predictive performance, sample-selection behavior, and sample efficiency. The final section summarizes the main findings and future research directions.

\section{Related work}
\label{sec:related_work}

\subsection{Data-driven modeling of HEA properties}
\label{subsec:hea_ml}

Materials informatics has enabled increasingly detailed prediction and screening of HEA properties. Rickman et al.~\cite{Rickman2019Informatics} demonstrated how statistical learning can organize multicomponent alloy spaces, while phase-prediction studies established efficient approaches for relating chemical descriptors to phase constitution~\cite{Huang2019Phase,Zhou2019PhaseDesign}. Property-focused investigations subsequently connected predictive models with experimental evaluation. Wen et al.~\cite{Wen2019HEAML} used ML and experimental feedback to identify HEAs with desirable hardness, and Bundela and Rahul~\cite{Bundela2022Mechanical} developed a microhardness-prediction framework with experimental validation. These studies demonstrate the value of data-driven relationships for guiding alloy development. For strength prediction, incorporating processing history is particularly important. Wang et al.~\cite{Wang2023HEAStrength} combined composition, processing conditions, and material descriptors in neural modeling of tensile properties, demonstrating experimentally validated alloy design. More recently, Alvi et al.~\cite{Alvi2025HEADeepGP} investigated prior-guided deep GPs for uncertainty-aware prediction of correlated HEA properties. Response representation, processing history, and predictive uncertainty are central to HEA strength modeling, motivating candidate selection that improves prediction under a restricted budget.

\subsection{Sequential learning in materials and manufacturing}
\label{subsec:materials_sl}

Sequential learning connects material-property prediction with adaptive experimental selection. Uncertainty-guided design studies have demonstrated how measured feedback can direct subsequent experiments toward useful material conditions~\cite{Balachandran2016Adaptive,Xue2016AdaptiveDesign,Lookman2019ActiveLearning}. Closed-loop investigations have extended this principle to autonomous materials characterization~\cite{Kusne2020ClosedLoop} and HEA discovery~\cite{Rao2022HEADiscovery}. For predictive modeling of material properties, Tian et al.~\cite{Tian2021PropertySurfaces} compared acquisition strategies across material-property curves and surfaces, showing that their effectiveness depends on the response characteristics, uncertainty estimates, noise, and available budget.

Related manufacturing studies have developed adaptive sampling strategies for learning complex process responses with limited observations. Mehta and Shao~\cite{Mehta2021AdaptiveSampling} developed variance-based adaptive sampling for multi-task GP modeling of manufactured surface shapes. Johnson et al.~\cite{Johnson2025ProjectionBO} combined GP regression with BO-guided experimentation to improve geometric accuracy in projection multi-photon 3D printing. Masinelli et al.~\cite{Masinelli2025ProcessingMaps} used uncertainty-driven experimentation and GP classification to construct laser powder bed fusion processing maps with fewer experiments. Raihan et al. demonstrated sample-efficient melt-pool geometry prediction and defect classification through surprise-guided learning and sequential random-forest sampling, respectively~\cite{Raihan2024Surprise,Raihan2025Defect}. Together, these studies demonstrate the value of adaptive sample selection for improving manufacturing predictions. For HEA strength prediction, we evaluate candidate samples by their reference-wide learning benefit and the robust predictive risk expected to remain.

\subsection{Acquisition strategies for sample-efficient prediction}
\label{subsec:gp_active_learning}

GP-based experimental design provides a tractable basis for evaluating how observations improve a predictive relationship. Classical active learning formulations consider the expected reduction in prediction error~\cite{Cohn1996AL}, while GP conditioning provides analytical expressions for updating predictions and uncertainty~\cite{RasmussenWilliams2006GPML}. EI and UCB offer established acquisition rules based on predicted responses and uncertainty~\cite{Jones1998EGO,Srinivas2010GPUCB}. Their candidate-wise scores, however, do not explicitly quantify the expected improvement in predictions across a reference domain. In the HEA strength prediction task, this broader influence is important because observations at related alloy conditions can provide overlapping information, while other conditions remain sparsely characterized.

Prediction-oriented acquisition explicitly considers this wider influence. Bickford Smith et al.~\cite{BickfordSmith2023EPIG} introduced expected predictive information gain, which evaluates possible outcomes through their information about target predictions. Its dependence on a specified target-input distribution makes the weighting of prediction conditions important. Riis et al.~\cite{Riis2022FullyBayesianGP} addressed uncertainty about the response model through fully Bayesian GP acquisition. Their mixture criterion captures predictive uncertainty and disagreement but ranks candidates using current mixture variance, leaving the expected reference-wide effect of an observation implicit. Distributionally robust acquisition considers alternative weightings of prediction conditions and protects against the largest associated risk~\cite{Frogner2021UnlabeledDRO}. Takeno et al.~\cite{Takeno2025DRAL} developed GP active-learning methods targeting worst-case expected prediction error. Their approximately greedy variance-reduction rule uses the current worst-case distribution during candidate ranking. However, a candidate's anticipated information can change which conditions dominate the remaining uncertainty, motivating a reassessment of these weights when evaluating its learning benefit.

For sample-efficient HEA strength modeling, these studies motivate a linked acquisition problem: anticipating how possible outcomes update the response predictions and the support for alternative covariance descriptions, while evaluating the robust uncertainty expected to remain across composition--processing conditions. ROSE-HEA addresses this challenge through an outcome-aware acquisition mechanism that estimates reference-wide learning gains and reoptimizes the worst-case reference weights for each candidate's expected residual risk. We evaluate the ROSE-HEA framework through matched-budget comparisons, selected-sample transfer, design-space coverage, regional accuracy, and budget sensitivity to assess its predictive performance, the broader value of acquired observations, and sample efficiency.
\section{Methodology}
\label{sec:methodology}

This section introduces the GP surrogate and conventional acquisition rules, develops the proposed outcome-aware robust risk-reduction policy, and describes the resulting ROSE-HEA workflow through the framework diagram and algorithm.

\subsection{Gaussian process surrogate model}
\label{subsec:rose_gp}

A GP provides a probabilistic representation of the input--response relationship, enabling strength prediction together with an estimate of uncertainty~\cite{RasmussenWilliams2006GPML}. Let $\boldsymbol{x}\in\mathbb{R}^{d}$ contain the $d$ composition and processing inputs, and let $y=f(\boldsymbol{x})+\varepsilon$ denote the measured UTS, where $f$ is the underlying response and $\varepsilon\sim\mathcal{N}(0,\tau^2)$ represents residual variation. At iteration $t$, the acquired observations are $\mathcal{D}_t=\{(\boldsymbol{x}_i,y_i):i\in I_t\}$, where $I_t$ indexes the $n_t$ acquired observations. The latent responses at these inputs satisfy
\begin{equation}
\boldsymbol{f}_t\mid\mathbf{X}_t\sim\mathcal{N}(\boldsymbol{m}_t,\mathbf{K}_t),
\label{eq:rose_gp_prior}
\end{equation}
where $\mathbf{X}_t\in\mathbb{R}^{n_t\times d}$ contains the acquired inputs as rows, $\boldsymbol{f}_t=[f(\boldsymbol{x}_i)]_{i\in I_t}$ and $\boldsymbol{m}_t=[m(\boldsymbol{x}_i)]_{i\in I_t}$ are column vectors in $\mathbb{R}^{n_t}$, and $\mathbf{K}_t\in\mathbb{R}^{n_t\times n_t}$ has entries $[\mathbf{K}_t]_{ij}=k(\boldsymbol{x}_i,\boldsymbol{x}_j)$. All vectors are treated as column vectors, with training quantities ordered consistently. The constant mean is estimated from the acquired responses. The covariance function describes how information is shared between input conditions. For the Mat\'ern $5/2$ covariance,
\begin{equation}
 k_{5/2}(\boldsymbol{x},\boldsymbol{x}')
 =\sigma_f^2\left(1+\sqrt{5}r+\frac{5r^2}{3}\right)e^{-\sqrt{5}r},
 \label{eq:rose_kernel}
\end{equation}
where $\sigma_f^2$ is the signal variance and $r^2=\sum_{k=1}^{d}(x_k-x_k')^2/\ell_k^2$ is the scaled squared distance. The correlation length $\ell_k$ controls sensitivity to input $k$. Covariance parameters and noise variance are fitted by maximizing the marginal likelihood. Conditioning on $\mathcal{D}_t$ gives the posterior mean and latent covariance:
\begin{equation}
\begin{aligned}
\mu_t(\boldsymbol{x})
&=m(\boldsymbol{x})
+\boldsymbol{k}_t(\boldsymbol{x})^{\mathsf T}
\mathbf{A}_t^{-1}(\boldsymbol{y}_t-\boldsymbol{m}_t),\\
c_t(\boldsymbol{x},\boldsymbol{x}')
&=k(\boldsymbol{x},\boldsymbol{x}')
-\boldsymbol{k}_t(\boldsymbol{x})^{\mathsf T}
\mathbf{A}_t^{-1}\boldsymbol{k}_t(\boldsymbol{x}'),
\end{aligned}
\end{equation}
where $\boldsymbol{k}_t(\boldsymbol{x}) =[k(\boldsymbol{x}_i,\boldsymbol{x})]_{i\in I_t} \in\mathbb{R}^{n_t}$, $\boldsymbol{y}_t\in\mathbb{R}^{n_t}$ contains the acquired responses, and $\mathbf{A}_t=\mathbf{K}_t+\tau^2\mathbf{I}_{n_t}$, with $\mathbf{I}_{n_t}$ denoting the identity matrix. The latent variance is $s_t^2(\boldsymbol{x})=c_t(\boldsymbol{x},\boldsymbol{x})$, and the variance of a future UTS outcome is $v_t(\boldsymbol{x})=s_t^2(\boldsymbol{x})+\tau^2$. These expressions are applied separately to the covariance hypotheses introduced below.

\subsection{Sequential selection and conventional acquisition rules}
\label{subsec:rose_sl_acquisition}

An acquisition function uses the current surrogate to rank the unlabeled candidates. EI and UCB provide conventional reference policies~\cite{Jones1998EGO,Srinivas2010GPUCB}:
\begin{equation}
\begin{aligned}
\alpha_{\mathrm{EI},t}(\boldsymbol{x})
&= \Delta_t(\boldsymbol{x})\Phi(z_t(\boldsymbol{x}))
+s_t(\boldsymbol{x})\phi(z_t(\boldsymbol{x})),\\
\alpha_{\mathrm{UCB},t}(\boldsymbol{x})
&= \mu_t(\boldsymbol{x})+\kappa s_t(\boldsymbol{x}),
\end{aligned}
\label{eq:rose_ei_ucb}
\end{equation}
where $\Delta_t(\boldsymbol{x})=\mu_t(\boldsymbol{x})-f_t^+-\xi$, $z_t(\boldsymbol{x})=\Delta_t(\boldsymbol{x})/s_t(\boldsymbol{x})$, and $f_t^+=\max_{i\in I_t}\mu_t(\boldsymbol{x}_i)$ is the best fitted response at an acquired input. The margin $\xi\geq0$ controls the required improvement, $\kappa\geq0$ controls the UCB uncertainty contribution, and $\Phi$ and $\phi$ are the standard normal distribution and density functions. When $s_t(\boldsymbol{x})=0$, EI takes its limiting value $\max\{\Delta_t(\boldsymbol{x}),0\}$.

\subsection{Proposed outcome-aware robust acquisition policy}
\label{subsec:rose_acquisition}

ROSE-HEA selects the next candidate condition by evaluating how its possible UTS outcomes could improve predictions across the composition--processing space~\cite{Cohn1996AL,BickfordSmith2023EPIG}. To compare candidates consistently, their expected learning benefits are evaluated over a fixed reference set $\mathcal{R}=\{\boldsymbol{r}_j\}_{j=1}^{M}$ containing $M$ input vectors. This set includes all development-row inputs, both acquired and unacquired, and excludes the test inputs. Its relationship to the initial set and candidate pool is illustrated in Fig.~\ref{fig:rose_framework}.

\subsubsection{Adaptive predictive representation}
\label{subsubsec:rose_adaptive_prediction}

The acquisition calculation begins by representing uncertainty about response smoothness using complementary Mat\'ern $5/2$, Mat\'ern $3/2$, and radial basis function (RBF) covariance hypotheses. Their predictions are combined through adaptive predictive-likelihood weights, allowing new observations to refine the support assigned to each response description~\cite{Riis2022FullyBayesianGP}:
\begin{equation}
 p_t(y\mid\boldsymbol{x})
 =\sum_{h=1}^{H}w_{h,t}\,
   \mathcal{N}\!\left(y;\mu_{h,t}(\boldsymbol{x}),v_{h,t}(\boldsymbol{x})\right),
 \label{eq:rose_predictive_mixture}
\end{equation}
where $h$ indexes the $H=3$ hypotheses, $w_{h,t}\geq0$ and $\sum_h w_{h,t}=1$ are their predictive weights, and $v_{h,t}(\boldsymbol{x}) =s_{h,t}^2(\boldsymbol{x})+\tau_h^2$ includes component-specific observation noise. Equal initial weights are updated using the predictive likelihood of each newly acquired response.

At a reference input $\boldsymbol{r}\in\mathcal{R}$, the latent mixture mean and variance are
\begin{equation}
\begin{aligned}
\overline{\mu}_t(\boldsymbol{r})
&= \sum_h w_{h,t}\mu_{h,t}(\boldsymbol{r}),\\
V_t(\boldsymbol{r})
&= \sum_h w_{h,t}\Big\{s_{h,t}^2(\boldsymbol{r})
+[\mu_{h,t}(\boldsymbol{r})
-\overline{\mu}_t(\boldsymbol{r})]^2\Big\},
\end{aligned}
\label{eq:rose_mixture_moments}
\end{equation}
where $\overline{\mu}_t$ is the weighted UTS prediction. The variance $V_t$ accounts for uncertainty within each hypothesis and disagreement between their predicted means. It provides the current reference-wise uncertainty against which the benefit of an additional observation is assessed.

\subsubsection{Expected learning from a candidate's UTS outcome}
\label{subsubsec:rose_learning_gain}

Once this predictive representation is established, the next step is to estimate how a candidate's UTS outcome could change predictions at the reference inputs. Possible outcomes at $\boldsymbol{x}$ are evaluated using Eq.~\eqref{eq:rose_predictive_mixture}. For an outcome $y$, conditioning each hypothesis gives
\begin{equation}
\mu_{h,t}^{\boldsymbol{x},y}(\boldsymbol{r})
=\mu_{h,t}(\boldsymbol{r})
+\frac{c_{h,t}(\boldsymbol{r},\boldsymbol{x})}
{v_{h,t}(\boldsymbol{x})}
[y-\mu_{h,t}(\boldsymbol{x})],
\label{eq:rose_hypothetical_gp_update}
\end{equation}
where $c_{h,t}(\boldsymbol{r},\boldsymbol{x})$ is the posterior covariance between the latent reference and candidate responses. The superscript $\boldsymbol{x},y$ denotes conditioning on the proposed observation. All fitted component parameters are held fixed during this prospective calculation.

The same outcome updates the predictive weights and resulting mixture mean:
\begin{equation}
\begin{aligned}
w_{h,t}^{\boldsymbol{x},y}
&=\frac{w_{h,t}L_{h,t}(y;\boldsymbol{x})}
{\sum_a w_{a,t}L_{a,t}(y;\boldsymbol{x})},\\
\overline{\mu}_t^{\boldsymbol{x},y}(\boldsymbol{r})
&=\sum_h w_{h,t}^{\boldsymbol{x},y}
\mu_{h,t}^{\boldsymbol{x},y}(\boldsymbol{r}),
\end{aligned}
\label{eq:rose_hypothetical_weights}
\end{equation}
where $L_{h,t}(y;\boldsymbol{x})=\mathcal{N}(y;\mu_{h,t}(\boldsymbol{x}),v_{h,t}(\boldsymbol{x}))$ is the component predictive likelihood and $a$ indexes the hypotheses in the normalizing sum. Observing a candidate's response can update both the conditional predictions and covariance-hypothesis support. To quantify the learning benefit, we average the squared changes in the reference predictions over the candidate's predictive distribution:
\begin{equation}
g_t(\boldsymbol{r},\boldsymbol{x})
=\mathbb{E}_{Y_{\boldsymbol{x}}}
\!\left[
\big(
\overline{\mu}_t^{\boldsymbol{x},Y_{\boldsymbol{x}}}
(\boldsymbol{r})
-\overline{\mu}_t(\boldsymbol{r})
\big)^2
\right],
\label{eq:rose_expected_gain}
\end{equation}
where $Y_{\boldsymbol{x}}$ is the unknown scalar outcome at $\boldsymbol{x}$, with predictive density $p_t(\cdot\mid\boldsymbol{x})$. By the law of total variance, $g_t$ equals the expected reduction in latent variance under the current fitted mixture. The expectation is evaluated by numerical quadrature, using the predictive distribution rather than the unobserved candidate response.

\subsubsection{Reference-wide robust risk reduction}
\label{subsubsec:rose_robust_risk}

After the expected gains have been obtained, the policy evaluates the robust risk that would remain after observing each candidate's response. Following distributionally robust active learning~\cite{Takeno2025DRAL}, the importance assigned to the reference inputs is allowed to vary around nominal probabilities $q_j=1/M$. The admissible weights and associated worst-case risk are
\begin{equation}
\begin{aligned}
\mathcal{P}_{\rho}(\boldsymbol{q}) &=\{\boldsymbol{p}\in\Delta_M: \|\boldsymbol{p}-\boldsymbol{q}\|_1\leq\rho\},\\ \mathcal{J}_{\rho}(\boldsymbol{v}) &=\max_{\boldsymbol{p}\in\mathcal{P}_{\rho}(\boldsymbol{q})} \boldsymbol{p}^{\mathsf T}\boldsymbol{v},
\end{aligned}
\label{eq:rose_robust_risk}
\end{equation}
where $\boldsymbol{q}=(q_1,\ldots,q_M)^{\mathsf T}$, with $q_j=1/M$, assigns equal nominal weight to each development row, $\Delta_M$ is the probability simplex, $\boldsymbol{p}$ is an admissible reference-weight vector, and $\boldsymbol{v}\in\mathbb{R}^{M}$ is a reference-wise risk vector. The radius $0\leq\rho\leq2$ limits the $L_1$ deviation from $\boldsymbol{q}$, allowing at most $\rho/2$ probability mass to be reassigned.

The proposed acquisition function, $\alpha_{\mathrm{ROSE},t}(\boldsymbol{x})$, measures the reduction in robust risk after accounting for the expected learning gain:
\begin{equation}
\boxed{
\alpha_{\mathrm{ROSE},t}(\boldsymbol{x})
= \mathcal{J}_{\rho}(\boldsymbol{V}_t)
- \mathcal{J}_{\rho}\!\left(
\boldsymbol{V}_t-\boldsymbol{g}_t(\boldsymbol{x})\right),
}
\label{eq:rose_acquisition}
\end{equation}
where $\boldsymbol{V}_t=[V_t(\boldsymbol{r}_j)]_{j=1}^{M}$ and $\boldsymbol{g}_t(\boldsymbol{x})=[g_t(\boldsymbol{r}_j,\boldsymbol{x})]_{j=1}^{M}$. The vector $\boldsymbol{V}_t-\boldsymbol{g}_t(\boldsymbol{x})$ represents the expected remaining latent variance. Possible outcomes are averaged first; the worst-case reference weights are then reoptimized separately for each candidate. Thus, the conditions dominating the residual risk can change according to the anticipated information provided by that candidate's response.

\subsection{ROSE-HEA sequential learning framework}
\label{subsec:rose_framework}

\begin{figure*}[!ht]
\centering
\includegraphics[width=0.85\textwidth]{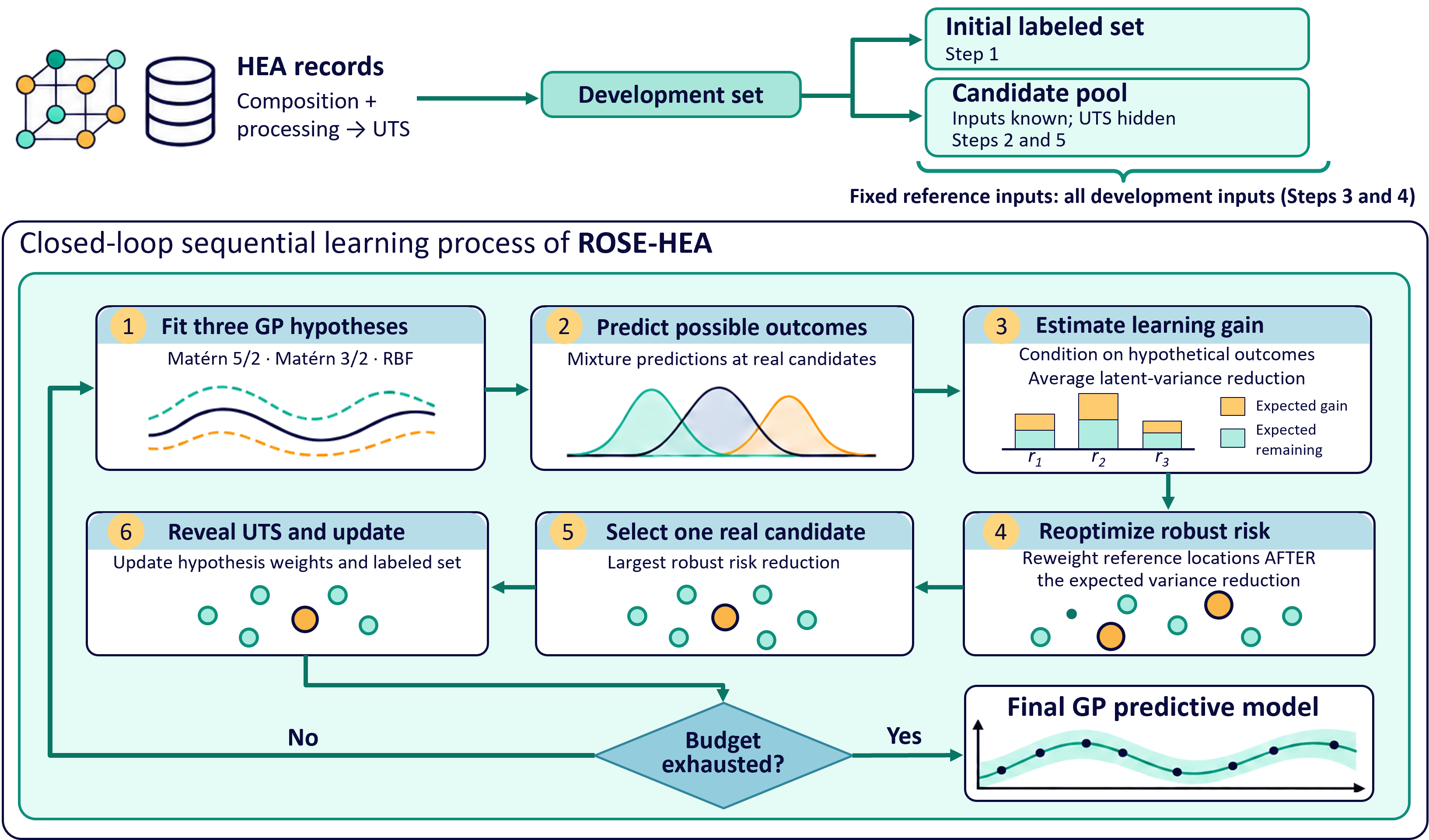}
\caption{ROSE-HEA framework showing data preparation, the six-step closed-loop acquisition process, and the final GP predictor. Reference inputs cover all development inputs and remain fixed during learning. The budget decision determines whether acquisition continues or the final predictor is returned.}
\label{fig:rose_framework}
\end{figure*}

Figure~\ref{fig:rose_framework} summarizes the data preparation, closed-loop acquisition process, and final predictive model of ROSE-HEA. Algorithm~\ref{alg:rose_framework} specifies the corresponding learning procedure. The development records are divided into an initial labeled set and a candidate pool with known inputs and unobserved UTS. Together, their input locations form the fixed reference set used to assess learning benefit.

The initial observations establish the covariance hypotheses and predictive mixture (Step~1). At each acquisition round, possible outcomes are predicted at the remaining candidates (Step~2). Hypothetical updates to the component means and weights yield the expected reference-wide learning gains in Eq.~\eqref{eq:rose_expected_gain} (Step~3), illustrated by the expected uncertainty reduction at reference locations. The worst-case reference weights are then reoptimized for each candidate's expected residual uncertainty using Eq.~\eqref{eq:rose_acquisition} (Step~4). These calculations use the current predictive distribution without revealing candidate responses. The candidate with the largest robust risk reduction is selected from the remaining pool (Step~5). After its UTS is observed (Step~6), the hypothesis weights are updated using the pre-update predictive likelihoods in Eq.~\eqref{eq:rose_hypothetical_weights}. The new observation is added to the training set, the selected candidate is removed from the pool, and the covariance hypotheses are refitted with the updated weights. The ``Budget exhausted?'' decision controls subsequent learning: the ``No'' branch returns to Step~1 for another acquisition round, whereas the ``Yes'' branch produces the final weighted GP predictor. The reference inputs remain fixed throughout this process, providing a consistent domain for evaluating successive learning gains.

\begin{algorithm}[!ht]
\caption{ROSE-HEA sequential learning for UTS prediction}
\label{alg:rose_framework}
\small

\begin{tabularx}{\linewidth}
{@{}l@{\enspace}>{\raggedright\arraybackslash}X@{}}
\textbf{Input:} &
Initial observations $\mathcal{D}_0$;
candidate indices $\mathcal{C}_0$ and inputs;
reference inputs $\mathcal{R}$;
sequential budget $B$;
ambiguity radius $\rho$.\\
\textbf{Output:} &
Final weighted UTS predictor and acquired observations.
\end{tabularx}

\begin{algorithmic}[1]

\StateWrap{Initialize equal hypothesis weights and fit the
covariance hypotheses on $\mathcal{D}_0$.}

\For{$t=0,\ldots,B-1$}

  \StateWrap{Calculate $\boldsymbol{V}_t$ using
  Eq.~\eqref{eq:rose_mixture_moments}.}

  \For{each candidate $i\in\mathcal{C}_t$}

    \StateWrap{Predict possible outcomes using
    Eq.~\eqref{eq:rose_predictive_mixture}.}

    \StateWrap{Apply hypothetical mean and weight updates using
    Eqs.~\eqref{eq:rose_hypothetical_gp_update}
    and \eqref{eq:rose_hypothetical_weights}.}

    \StateWrap{Average over the outcomes to obtain
    $\boldsymbol{g}_t(\boldsymbol{x}_i)$ from
    Eq.~\eqref{eq:rose_expected_gain}.}

    \StateWrap{Reoptimize the expected residual reference risk
    and calculate $\alpha_{\mathrm{ROSE},t}(\boldsymbol{x}_i)$
    using Eq.~\eqref{eq:rose_acquisition}.}

  \EndFor

  \StateWrap{Select
  $i_{t+1}\in\arg\max_{i\in\mathcal{C}_t}
  \alpha_{\mathrm{ROSE},t}(\boldsymbol{x}_i)$
  and observe $y_{i_{t+1}}$.}

  \StateWrap{Update hypothesis weights using the pre-update
  predictive likelihoods in
  Eq.~\eqref{eq:rose_hypothetical_weights}.}

  \StateWrap{Set
  $\mathcal{D}_{t+1}
  =\mathcal{D}_t\cup
  \{(\boldsymbol{x}_{i_{t+1}},y_{i_{t+1}})\}$
  and
  $\mathcal{C}_{t+1}
  =\mathcal{C}_t\setminus\{i_{t+1}\}$.}

  \StateWrap{Refit the covariance hypotheses on
  $\mathcal{D}_{t+1}$, retaining the updated weights.}

\EndFor

\StateWrap{\Return the final weighted UTS predictor
and acquired observations.}

\end{algorithmic}
\end{algorithm}

\section{Dataset and experimental protocol}
\label{sec:dataset_experimental_protocol}

This section describes the HEA data and the experimental protocol used to assess predictive performance under a limited training budget. It introduces the comparison methods, explains how the data are partitioned and training observations are allocated, and specifies the model-training procedures and prediction metrics.

\subsection{HEA dataset and input representation}
\label{subsec:hea_dataset_representation}

The study uses the tensile-property database reported by Wang et al.~\cite{Wang2023HEAStrength}, comprising 501 records described by 14 composition and processing inputs. Ten variables represent the atomic percentages of C, Al, V, Cr, Mn, Fe, Co, Ni, Cu, and Mo. The remaining four describe homogenization temperature, cold-rolling thickness reduction, annealing temperature, and annealing duration. The target is UTS, ranging from 320 to 1867~MPa, with a mean of 895.36~MPa and a standard deviation of 305.46~MPa. Table~\ref{tab:hea_dataset_variables} summarizes the variables and their recorded ranges.

\begin{table}[!ht]
\centering
\caption{Inputs and target response for the 501 HEA records. Ranges reflect recorded values; processing zeros are dataset encodings.}
\label{tab:hea_dataset_variables}
\small
\begin{tabularx}{\columnwidth}{@{}lXr@{}}
\toprule
\shortstack[l]{Variable\\group} &
\shortstack[l]{Modeled variable\\(unit)} &
\shortstack[r]{Recorded\\range} \\
\midrule
Composition & Carbon, C (at.\%) & 0--6 \\
 & Aluminum, Al (at.\%) & 0--16.67 \\
 & Vanadium, V (at.\%) & 0--33.3333 \\
 & Chromium, Cr (at.\%) & 0--39.4 \\
 & Manganese, Mn (at.\%) & 0--50 \\
 & Iron, Fe (at.\%) & 0--59 \\
 & Cobalt, Co (at.\%) & 0--50 \\
 & Nickel, Ni (at.\%) & 0--60 \\
 & Copper, Cu (at.\%) & 0--36 \\
 & Molybdenum, Mo (at.\%) & 0--10 \\
\midrule
Processing & Homogenization temperature (K) & 0--1573 \\
 & Cold-rolling thickness reduction (\%) & 0--96 \\
 & Annealing temperature (K) & 0--1523 \\
 & Annealing duration (h) & 0--50 \\
\midrule
Response & Ultimate tensile strength (MPa) & 320--1867 \\
\bottomrule
\end{tabularx}
\end{table}

\subsection{Comparison methods and sampling protocol}
\label{subsec:hea_benchmarks}
\label{subsec:hea_matched_holdouts}

The main comparison includes four sequential strategies: ROSE-HEA, EI-BO, UCB-BO, and Random GP. ROSE-HEA, EI-BO, and UCB-BO select candidates using the acquisition policies described in Section~\ref{sec:methodology}; Random GP combines a GP predictor with uniform random selection. Eight conventional regressors provide additional benchmarks: Extra Trees, Lasso, Linear Regression, a multilayer perceptron (MLP), Random Forest, radial-basis-function support vector regression (RBF-SVR), Ridge, and XGBoost.

Thirty repeated runs assess variation arising from data partitioning and initial-sample selection. In each run, approximately 80\% of the records form the development set and 20\% form the test set. The development set comprises the non-test records available for initialization and subsequent sample selection. Records with identical composition and processing inputs are kept in the same partition, giving 398--401 development records and 100--103 test records. All methods share the same test set within a run, while the partition is resampled between runs.

From the development set, 25 records are randomly selected to form the common initial training set. The remaining 373--376 records form the candidate pool. Candidate input values are available during selection, while their recorded UTS values are revealed only when selected. For ROSE-HEA, all development-row input vectors also form the fixed reference set $\mathcal{R}$ introduced in Section~\ref{subsec:rose_acquisition}. These inputs provide the locations at which anticipated learning gains are evaluated. The reference set retains both acquired and unacquired inputs throughout a run, while the candidate pool shrinks as samples are selected.

Each sequential strategy selects 100 additional candidates, revealing one UTS value at a time, to obtain 125 training observations. An observation consists of a record's inputs and revealed UTS value. The eight conventional regressors share the 25 initial observations and 100 random additions with Random GP. All methods therefore receive 125 training observations and use the same test set within a run. Test inputs and responses are excluded from the reference set, acquisition, model fitting, and parameter selection.

\subsection{Model configuration and training}
\label{subsec:hea_preprocessing_tuning}

With the training-budget design established, the models are configured and fitted using only the available training observations. The GP benchmarks use Mat\'ern $5/2$ covariance, while ROSE-HEA uses the adaptive predictive representation developed in Section~\ref{sec:methodology}. EI and UCB follow the stated acquisition definitions~\cite{Jones1998EGO,Srinivas2010GPUCB}, with $\xi=0.01$ and $\kappa=2$, respectively; ROSE-HEA uses $\rho=0.50$. GP parameters are fitted by marginal likelihood, with common fitting settings and parameter bounds for shared covariance families~\cite{RasmussenWilliams2006GPML}. Inputs and UTS are standardized using the common initial training set, and this scaling is retained throughout each run.

For the conventional regressors, the tree ensembles use 300 estimators, and the MLP has hidden layers of 64, 32, and 16 units with L-BFGS training. Tunable parameters are selected through three-fold cross-validation on the 125 available training observations, keeping identical input conditions within the same fold. Up to 12 parameter configurations are considered per model; Linear Regression requires no hyperparameter search. Input and target scaling are fitted within each training fold. The selected configuration is then refitted on all 125 observations. Test data are excluded from preprocessing and tuning, and predictions are expressed in MPa for evaluation.

\subsection{Prediction metrics}
\label{subsec:hea_metrics_reporting}

Predictive accuracy is assessed using root-mean-square error (RMSE), mean absolute error (MAE), and the coefficient of determination ($R^2$):
\begin{align}
\mathrm{RMSE} &= \sqrt{\frac{1}{N}\sum_{i=1}^{N}e_i^2},
\quad
\mathrm{MAE} = \frac{1}{N}\sum_{i=1}^{N}|e_i|,
\label{eq:hea_rmse_mae}\\
R^2 &= 1-\frac{\sum_{i=1}^{N}e_i^2}
{\sum_{i=1}^{N}(y_i-\overline{y})^2},
\label{eq:hea_r2}
\end{align}
where $N$ is the number of test records, $e_i=y_i-\widehat{y}_i$ is the prediction error, $y_i$ and $\widehat{y}_i$ are the observed and predicted UTS, and $\overline{y}$ is the mean test response. RMSE and MAE are expressed in MPa, while $R^2$ is dimensionless. Metrics are calculated separately for each run and summarized by their mean and standard deviation across the 30 runs. Comparisons are paired within runs, and percentage error reductions are calculated relative to the reference method's mean error.

\section{Results and discussion}
\label{sec:results_discussion}

The results assess how effectively ROSE-HEA converts a limited training budget into accurate UTS predictions for HEAs. The analysis progresses from final-budget performance and sequential learning behavior to the predictive value and design-space distribution of the selected observations. Budget studies then examine how initialization, sequential acquisition, and total training size influence accuracy and sample efficiency. Unless stated otherwise, numerical summaries report results across 30 matched runs, with percentage error reductions calculated from the across-run means.

\subsection{Comparative predictive performance at a fixed training budget}
\label{subsec:results_final_performance}

\begin{figure*}[!ht]
\centering
\includegraphics[width=0.90\textwidth]{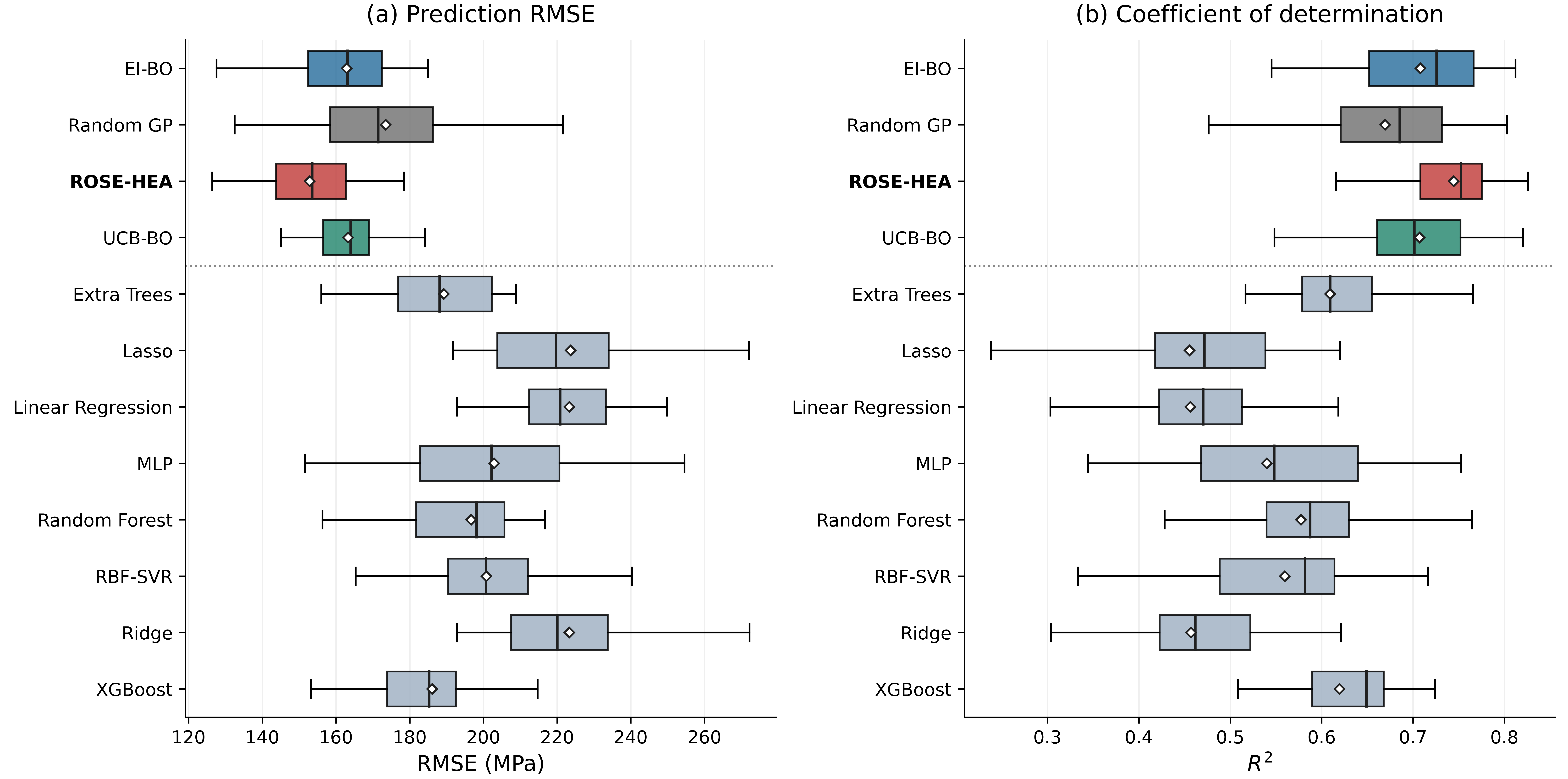}
\caption{Prediction at 125 training observations across 30 runs: (a) RMSE and (b) $R^2$. Boxes show interquartile ranges, lines indicate medians, diamonds denote means, and whiskers follow the 1.5-IQR rule. The dotted separator distinguishes SL and conventional ML methods.}
\label{fig:rose_final_boxplots}
\end{figure*}

We first compare the proposed ROSE-HEA framework with EI-BO, UCB-BO, and Random GP under a common budget of 125 training observations. ROSE-HEA achieves the lowest mean RMSE and highest mean $R^2$ among these sequential approaches (Fig.~\ref{fig:rose_final_boxplots}). Its mean RMSE is 152.84~MPa, compared with 162.88~MPa for EI-BO, 163.25~MPa for UCB-BO, and 173.46~MPa for Random GP, corresponding to reductions of 6.16\%, 6.37\%, and 11.89\%, respectively. ROSE-HEA also attains a mean $R^2$ of 0.7443, compared with 0.7080, 0.7070, and 0.6693 for these benchmarks.

The advantage also extends to conventional regression. XGBoost, the strongest conventional predictor at this budget, attains a mean RMSE of 186.07~MPa and $R^2$ of 0.6195; ROSE-HEA reduces its mean RMSE by 17.86\%. The distributions further show that the improvement recurs across changing initial samples and held-out partitions. ROSE-HEA achieves lower RMSE than EI-BO in 23 of 30 paired runs, UCB-BO in 21 runs, and Random GP in 25 runs. Against the four strongest conventional regressors by mean RMSE---XGBoost, Extra Trees, Random Forest, and RBF-SVR---ROSE-HEA achieves lower RMSE in 29 of 30 paired runs for each comparison. These results demonstrate a recurring predictive advantage across both sequential and conventional learning benchmarks under the common training budget.

The MAE comparison provides a complementary assessment of absolute prediction errors (Table~\ref{tab:rose_final_performance}). ROSE-HEA attains the lowest mean MAE, at $115.18\pm10.23$~MPa. Relative reductions range from 7.49\% to 9.71\% against the GP benchmarks and from 18.01\% to 35.11\% against the conventional regressors. In particular, the 18.01\% reduction relative to XGBoost confirms an advantage over the strongest conventional model, alongside the larger improvements over less accurate predictors. The agreement across RMSE, MAE, and $R^2$ supports the framework's effectiveness in predicting UTS with limited observations.

\begin{table}[!ht]
\centering
\caption{MAE at 125 observations (mean $\pm$ SD across
30 runs) and ROSE-HEA's reduction relative to each benchmark.}
\label{tab:rose_final_performance}
\small
\begin{tabular*}{\columnwidth}{@{\extracolsep{\fill}}lrr@{}}
\toprule
\textbf{Method} &
\shortstack[r]{\textbf{MAE}\\(MPa)} &
\shortstack[r]{\textbf{ROSE-HEA}\\\textbf{reduction (\%)}} \\
\midrule
\textbf{ROSE-HEA} & $\mathbf{115.18}\pm10.23$ & --- \\
EI-BO             & $125.20\pm14.71$ & 8.00 \\
Random GP         & $127.56\pm13.24$ & 9.71 \\
UCB-BO            & $124.51\pm9.15$  & 7.49 \\
\midrule
Extra Trees       & $140.73\pm13.77$ & 18.16 \\
Lasso             & $177.49\pm18.98$ & 35.11 \\
Linear Regression & $175.92\pm16.10$ & 34.53 \\
MLP               & $140.90\pm16.36$ & 18.26 \\
Random Forest     & $150.10\pm14.06$ & 23.26 \\
RBF-SVR           & $149.88\pm15.25$ & 23.15 \\
Ridge             & $177.17\pm16.49$ & 34.99 \\
XGBoost           & $140.49\pm14.05$ & 18.01 \\
\bottomrule
\end{tabular*}
\end{table}

\subsection{Sequential learning progression}
\label{subsec:results_learning_progression}

\begin{figure*}[!ht]
\centering
\includegraphics[width=0.90\textwidth]{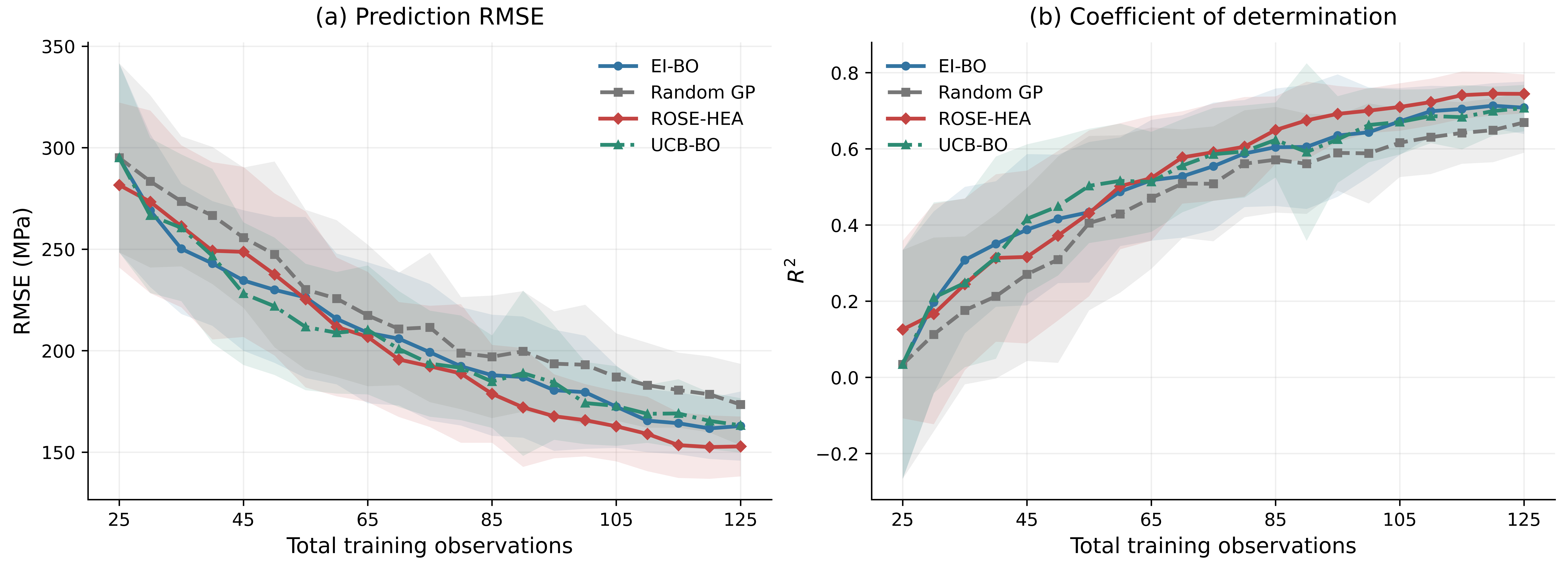}
\caption{Sequential learning from 25 initial to 125 total observations: (a) RMSE and (b) $R^2$. Lines show means and shaded bands show $\pm$ one SD across 30 matched runs.}
\label{fig:rose_learning_curves}
\end{figure*}

Learning progression reveals how rapidly each framework turns additional observations into improved predictions. The four sequential approaches begin with the same 25 observations and acquire 100 additional observations. Each retains its prescribed predictive representation, so the trajectories capture the combined effects of prediction and sample selection under common initial information. The resulting RMSE and $R^2$ curves are shown in Fig.~\ref{fig:rose_learning_curves}.

ROSE-HEA's comparative advantage becomes more pronounced as acquired information accumulates. At 50 observations, its mean RMSE is 237.58~MPa, compared with 229.98~MPa for EI-BO and 221.90~MPa for UCB-BO. By 75 observations, it reaches 192.29~MPa, compared with 199.23 and 193.62~MPa, respectively. At 100 observations, ROSE-HEA attains 165.76~MPa, against 179.54~MPa for EI-BO and 174.21~MPa for UCB-BO. Thus, the final accuracy advantage develops through stronger later-stage learning, with the relative ranking changing during acquisition.

This pattern is consistent with the reference-wide acquisition objective in Eq.~\eqref{eq:rose_acquisition}. Candidates are valued by the expected influence of their UTS outcomes on predictions, covariance-hypothesis support, and residual robust risk across related conditions. As the response representation is refined, subsequent acquisitions can be directed using a better-informed assessment of these learning benefits. The simultaneous improvements in RMSE and $R^2$ indicate that the acquired observations enhance UTS prediction across alloy conditions.

\subsection{Predictive value of ROSE-HEA-selected observations}
\label{subsec:results_sample_transfer}

\begin{figure*}[!ht]
\centering
\includegraphics[width=0.90\textwidth]{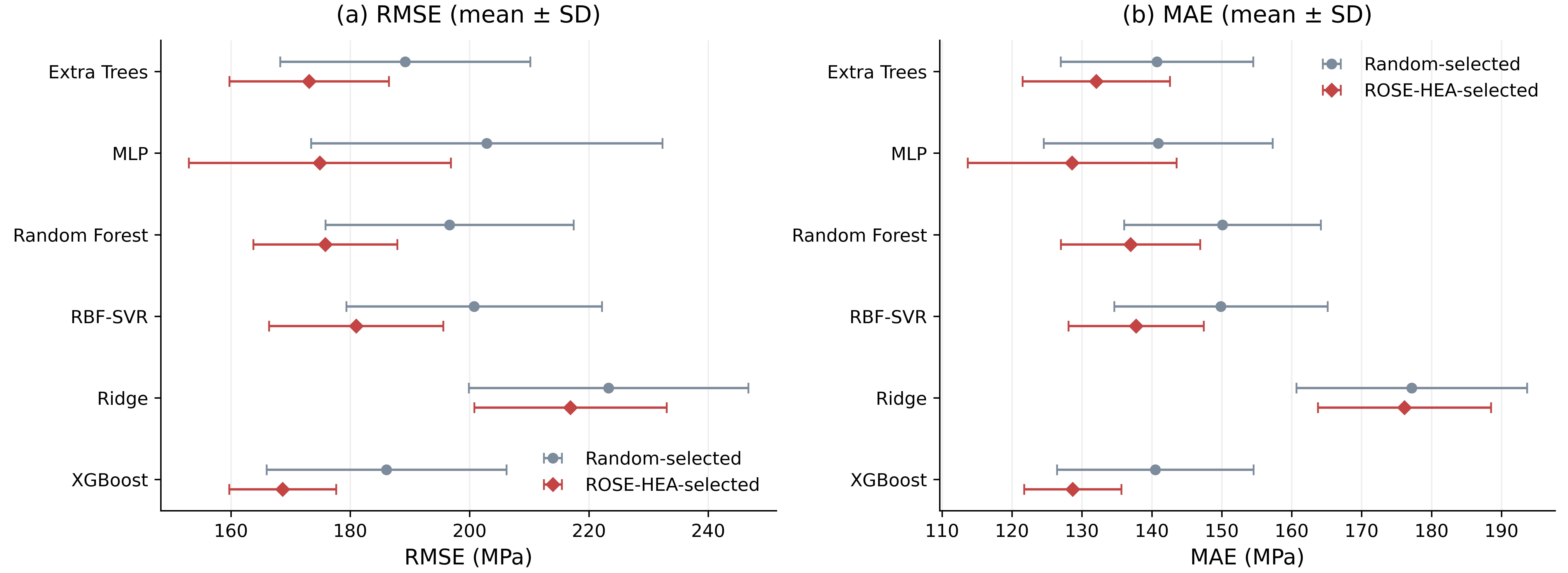}
\caption{Prediction using ROSE-HEA-selected and randomly selected observations: (a) RMSE and (b) MAE. Each pair uses 125 training observations and a common test set. Markers and error bars show means and $\pm$ one SD across 30 matched runs.}
\label{fig:rose_sample_transfer}
\end{figure*}

An informative training set should support accurate prediction beyond the particular model that generated its selection trajectory. To assess this property, six ML regressors are fitted separately to ROSE-HEA-selected and randomly selected training sets. Each pair contains the common 25 initial observations and 100 subsequent selections, and uses the same held-out partition and model-selection procedure described in Section~\ref{subsec:hea_preprocessing_tuning}. Hyperparameters are selected independently for each training set using training-only cross-validation. This comparison evaluates whether the selected observations remain useful across different predictor families at an equal training budget. All six regressors achieve lower mean RMSE with ROSE-HEA-selected observations (Fig.~\ref{fig:rose_sample_transfer}(a)). The reductions are 8.52\% for Extra Trees, 13.80\% for MLP, 10.59\% for Random Forest, 9.85\% for RBF-SVR, 2.86\% for Ridge, and 9.35\% for XGBoost. For MLP, mean RMSE decreases from 202.88 to 174.89~MPa, while XGBoost improves from 186.07 to 168.67~MPa.

Mean MAE also decreases for all six regressors (Fig.~\ref{fig:rose_sample_transfer}(b)). It falls from 140.90 to 128.59~MPa for MLP and from 150.10 to 136.95~MPa for Random Forest. Ridge shows a smaller improvement, from 177.17 to 176.12~MPa. The agreement between RMSE and MAE indicates that the benefits extend to both larger prediction errors and average absolute errors. The larger gains for several nonlinear regressors suggest that their flexibility helps exploit the information in the selected observations. Although model flexibility influences the magnitude of these gains, improvements across all six regressors demonstrate that ROSE-HEA's acquisition decisions produce useful training sets beyond their use within the framework's own predictor.

\begin{figure*}[!ht]
\centering
\includegraphics[width=\textwidth]{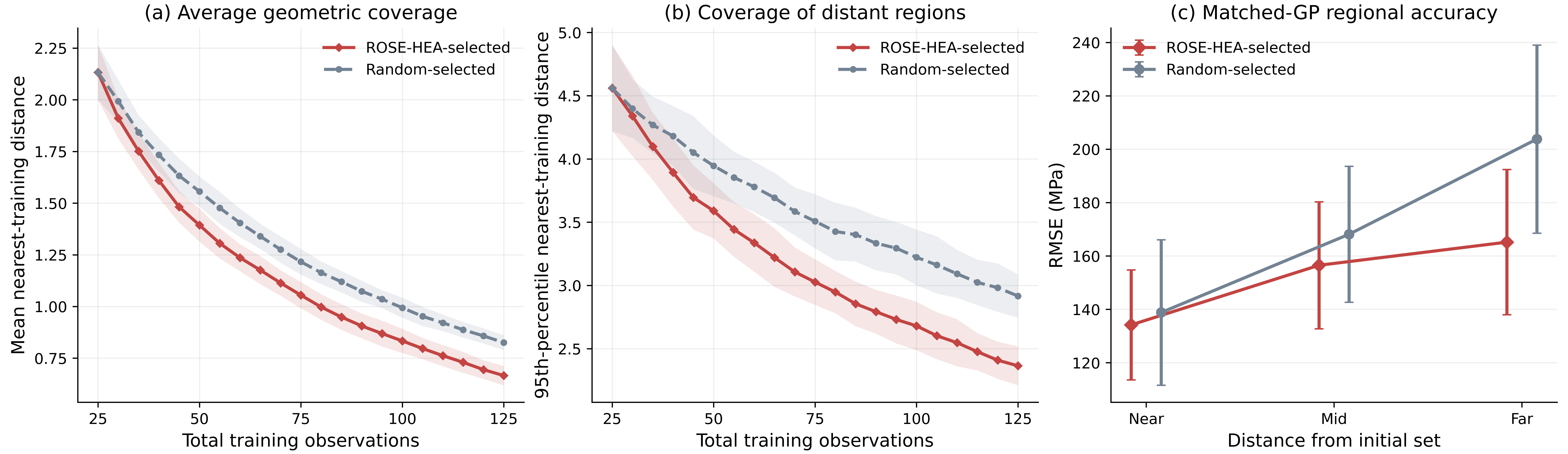}
\caption{Coverage and regional accuracy:
(a) mean and (b) 95th-percentile nearest-training distance;
(c) regional RMSE using a common Mat\'ern $5/2$ GP.
Lines show means; bands and error bars show $\pm$ one SD across 30 runs.}
\label{fig:rose_coverage_regions}
\end{figure*}

\subsection{Design-space coverage and regional prediction accuracy}
\label{subsec:results_coverage_regions}

We examine how the selected observations represent the composition--processing domain and support prediction at initially distant conditions. Coverage is measured by the distance from each fixed development input, including acquired rows, to its nearest training sample in the full 14-dimensional space. Development-set standardization is used for this geometric analysis, separately from GP preprocessing. Lower mean and 95th-percentile distances indicate better average coverage and fewer poorly represented conditions. At 125 observations, ROSE-HEA reduces mean nearest-training distance from 0.8255 to 0.6658, a 19.35\% improvement, and the 95th-percentile distance from 2.9166 to 2.3648, an 18.92\% improvement (Fig.~\ref{fig:rose_coverage_regions}(a,b)). Both measures are lower than random selection in all 30 runs, demonstrating a consistent coverage advantage across the full input space.

\begin{figure*}[!ht]
\centering
\includegraphics[width=0.90\textwidth]{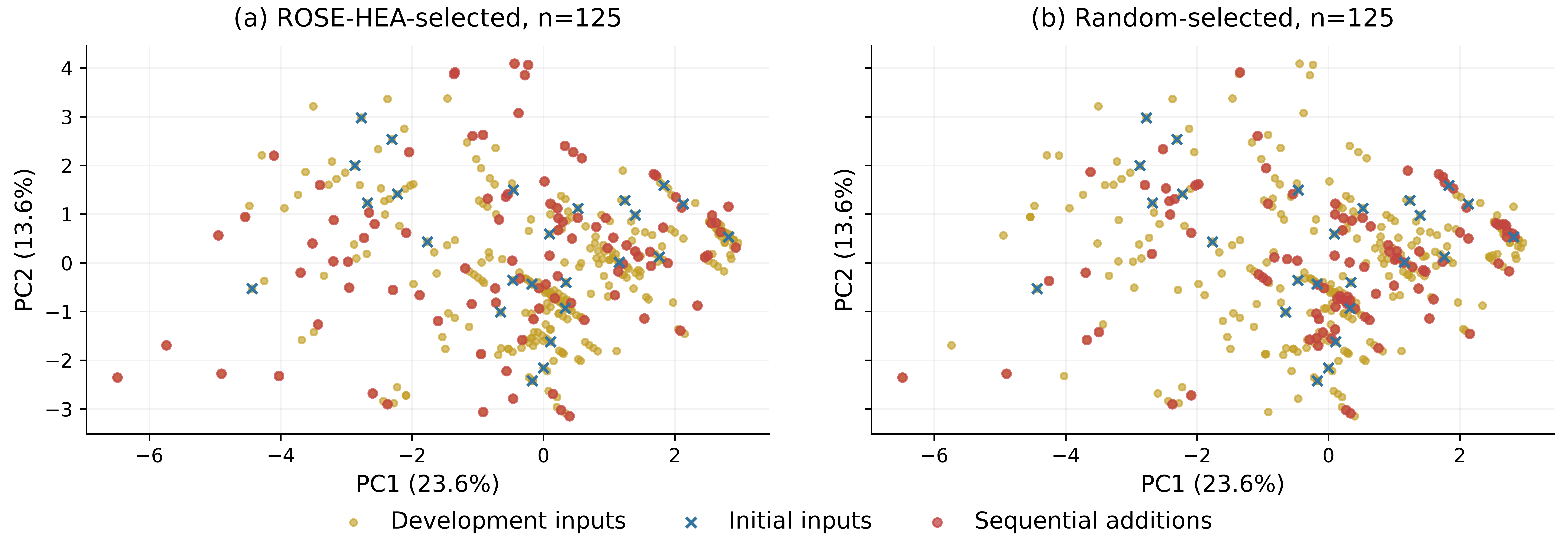}
\caption{PCA projection of 125 observations in run~1:
(a) ROSE-HEA selection and (b) random selection.
Gold points show development inputs, blue crosses the 25 initial inputs,
and red points the 100 subsequent selections.}
\label{fig:rose_pca_selection}
\end{figure*}

Regional analysis examines whether this coverage advantage is accompanied by improved prediction at initially distant conditions. Test inputs are divided into near, intermediate, and distant groups according to thirds of their distances from the 25 initial inputs. The groups are identical for both selection strategies within each run. A common Mat\'ern $5/2$ GP, using the same fitting and normalization settings, is fitted separately to each selected training set. ROSE-HEA-selected observations reduce mean regional RMSE by 3.35\%, 6.91\%, and 18.95\%, respectively (Fig.~\ref{fig:rose_coverage_regions}(c)). The largest improvement occurs in the distant group, where RMSE decreases from 203.81 to 165.19~MPa and is lower in 25 of 30 paired runs. Distant-group MAE also decreases by 18.00\%. From these results, we find that broader coverage accompanies improved UTS prediction at conditions with limited initial geometric support.

The PCA projection illustrates differences in the sampling patterns (Fig.~\ref{fig:rose_pca_selection}). ROSE-HEA distributes selections across separated and peripheral locations, while random selection places more observations within some densely populated clusters. Because random selection assigns equal probability to each candidate record, regions containing more records are more likely to receive multiple observations. The projection captures approximately 37.2\% of input variation and provides a visual complement to the full-space analysis. Broader coverage and improved regional accuracy support the interpretation that ROSE-HEA selects informative candidates beyond densely sampled neighborhoods.

\subsection{Sensitivity to training budget and comparative sample efficiency}
\label{subsec:results_budget_efficiency}

\subsubsection{Initial-size and sequential-budget sensitivity}
\label{subsubsec:results_budget_sensitivity}

\begin{figure*}[!ht]
\centering
\includegraphics[width=0.90\textwidth]{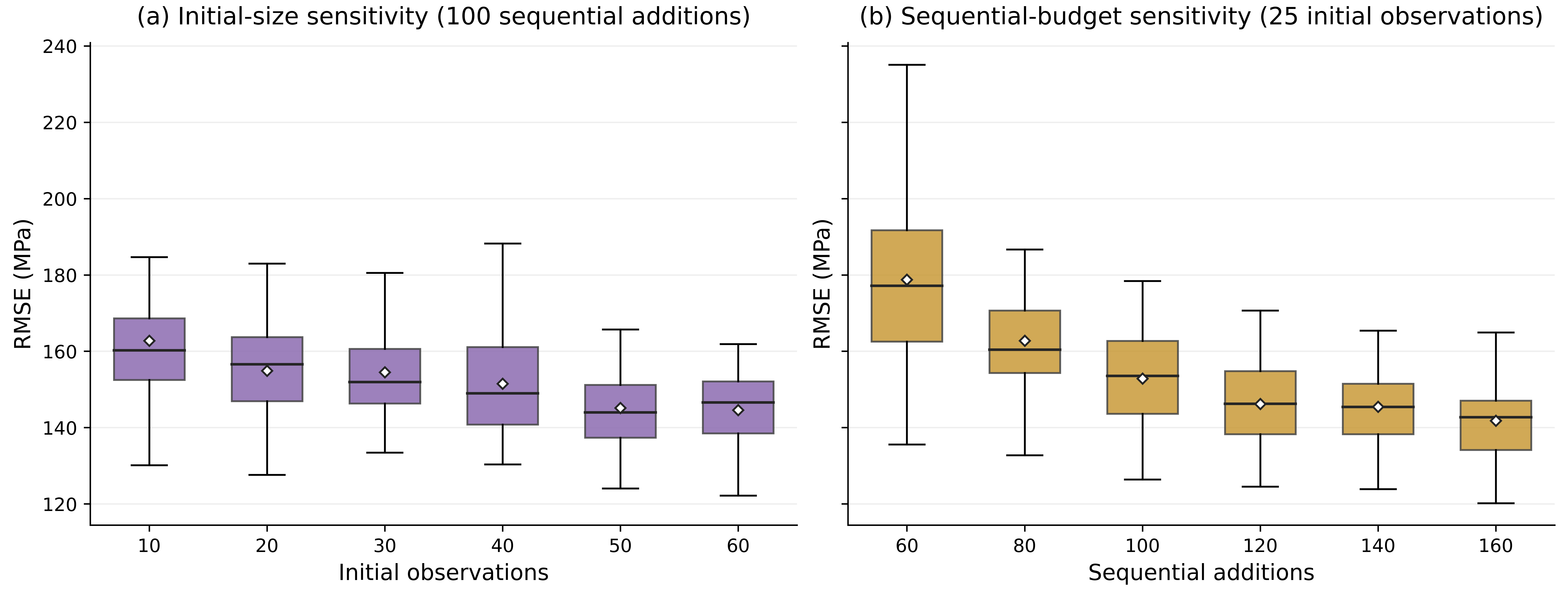}
\caption{ROSE-HEA budget sensitivity across 30 matched runs: (a) 10--60 initial observations with 100 sequential additions; (b) 60--160 additions after 25 initial observations. Total budgets are 110--160 and 85--185, respectively. Boxplot conventions follow Fig.~\ref{fig:rose_final_boxplots}.}
\label{fig:rose_budget_boxplots}
\end{figure*}

Budget sensitivity examines how prediction accuracy depends on the amount of initial information and the opportunity for subsequent adaptive selection. The initial-size study uses nested sets of 10, 20, 30, 40, 50, and 60 observations within each run, with a common test partition and preprocessing fitted to each initialization. Each initial set is followed by its own 100-step acquisition trajectory, giving total budgets of 110--160 observations. Increasing the initial size from 10 to 60 lowers mean RMSE from 162.77 to 144.60~MPa and increases mean $R^2$ from 0.7097 to 0.7708 (Fig.~\ref{fig:rose_budget_boxplots}(a)). The 11.16\% RMSE reduction reflects the combined benefit of a larger starting set and a larger total budget.

The sequential-budget study fixes initialization at 25 observations and varies the additions across 60, 80, 100, 120, 140, and 160, giving total budgets of 85--185 observations. Within each run, the settings are nested prefixes of one acquisition trajectory and use the same test set. Increasing the additions from 60 to 160 lowers mean RMSE from 178.77 to 141.82~MPa, a 20.67\% reduction, and raises mean $R^2$ from 0.6493 to 0.7793 (Fig.~\ref{fig:rose_budget_boxplots}(b)). The distributions shift toward lower errors as the training budget increases, although incremental gains vary across budget intervals. Within the tested ranges, increasing the randomly selected initial set produces a comparatively modest accuracy gain, whereas extending sequential acquisition yields a larger overall improvement. ROSE-HEA can therefore refine UTS predictions from a small number of initial observations by directing the sequential budget toward candidates with high expected learning value.

\begin{table*}[!ht]
\centering
\caption{RMSE (MPa; mean $\pm$ SD) across 30 matched runs.
Conventional regressors use nested random training sets;
ROSE-HEA uses sequential selection. Bold entries indicate
the lowest mean at each budget.}
\label{tab:rose_budget_comparison}
\small
\setlength{\tabcolsep}{3pt}
\begin{tabular*}{\textwidth}{@{\extracolsep{\fill}}lrrrrrr@{}}
\toprule
\textbf{Method} &
\shortstack{\textbf{85}\\\textbf{observations}} &
\shortstack{\textbf{105}\\\textbf{observations}} &
\shortstack{\textbf{125}\\\textbf{observations}} &
\shortstack{\textbf{145}\\\textbf{observations}} &
\shortstack{\textbf{165}\\\textbf{observations}} &
\shortstack{\textbf{185}\\\textbf{observations}} \\
\midrule
\textbf{ROSE-HEA}
& $\mathbf{178.77}\pm24.12$
& $\mathbf{162.75}\pm17.26$
& $\mathbf{152.84}\pm14.74$
& $\mathbf{146.18}\pm12.32$
& $\mathbf{145.47}\pm11.76$
& $\mathbf{141.82}\pm10.91$ \\
Extra Trees
& $205.83\pm22.70$ & $194.88\pm20.66$ & $189.22\pm20.96$
& $181.01\pm20.99$ & $174.46\pm21.15$ & $168.20\pm19.02$ \\
Lasso
& $237.21\pm27.88$ & $225.36\pm26.95$ & $223.64\pm26.62$
& $218.20\pm17.74$ & $215.71\pm18.01$ & $214.29\pm17.03$ \\
Linear Regression
& $255.57\pm97.19$ & $232.02\pm41.49$ & $223.29\pm22.36$
& $236.96\pm103.42$ & $223.79\pm49.82$ & $221.35\pm46.29$ \\
MLP
& $214.76\pm31.54$ & $208.15\pm47.73$ & $202.88\pm29.43$
& $191.20\pm30.97$ & $184.30\pm31.06$ & $183.44\pm35.27$ \\
Random Forest
& $211.77\pm20.53$ & $204.26\pm19.84$ & $196.63\pm20.80$
& $188.63\pm21.24$ & $179.10\pm21.40$ & $173.42\pm19.67$ \\
RBF-SVR
& $222.53\pm26.23$ & $210.33\pm24.91$ & $200.76\pm21.43$
& $196.09\pm22.90$ & $188.61\pm22.09$ & $186.85\pm21.98$ \\
Ridge
& $237.90\pm28.89$ & $226.51\pm24.38$ & $223.29\pm23.42$
& $218.80\pm17.07$ & $216.90\pm17.83$ & $214.41\pm16.93$ \\
XGBoost
& $198.88\pm19.92$ & $189.96\pm20.84$ & $186.07\pm20.11$
& $174.48\pm20.51$ & $168.77\pm20.81$ & $161.90\pm19.77$ \\
\bottomrule
\end{tabular*}
\end{table*}

\subsubsection{Sample efficiency relative to conventional regression}
\label{subsubsec:results_ml_measurement_efficiency}

We further compare ROSE-HEA with the eight ML regressors using 85, 105, 125, 145, 165, and 185 training observations. Within each run, the regressors share nested random training sets containing the common 25 initial observations and use the same test partition as ROSE-HEA. Training-only tuning is repeated at each budget following Section~\ref{subsec:hea_preprocessing_tuning}, while ROSE-HEA uses its sequentially selected observations. ROSE-HEA achieves the lowest mean RMSE at every budget, with XGBoost remaining the strongest conventional regressor (Table~\ref{tab:rose_budget_comparison}). This advantage persists as additional training observations become available.

The unequal-budget comparisons demonstrate the framework's sample efficiency. With 125 observations, ROSE-HEA achieves a lower mean RMSE than XGBoost trained on 185 observations (152.84 versus 161.90~MPa). It therefore outperforms the strongest tested conventional regressor while using 60 fewer observations, corresponding to a 32.4\% reduction in the training budget. Similarly, ROSE-HEA achieves a mean RMSE of 162.75~MPa with 105 observations, compared with 168.77~MPa for XGBoost trained on 165 observations. This comparison shows better mean prediction accuracy with 60 fewer observations, or a 36.4\% reduction in the reference training budget. Across the evaluated budgets from 105 to 185 observations, ROSE-HEA's across-run RMSE standard deviation decreases steadily from 17.26 to 10.91~MPa while remaining lower than that of all eight conventional regressors, indicating increasingly consistent predictive performance as sequential observations are added.
\section{Conclusion}
\label{sec:conclusion}

This study developed ROSE-HEA for sample-efficient prediction of ultimate tensile strength in high-entropy alloys using composition and processing information. Its outcome-aware acquisition evaluates candidates through their expected contribution to predictions across relevant alloy conditions. Adaptive predictive weighting and reassessment of residual robust risk provide a unified mechanism for progressively improving strength prediction. Across 30 matched runs, ROSE-HEA achieved the strongest mean predictive performance among the evaluated sequential and conventional learning approaches, including a 6.16\% reduction in mean RMSE relative to the well-established EI-BO benchmark at the common training budget. The selected observations also improved prediction across all six regressors examined in the transfer analysis, demonstrating their usefulness beyond ROSE-HEA's own predictor. In the full 14-dimensional design space, mean nearest-training distance was reduced by 19.35\% relative to random selection, with improvement in every run. Together with the larger prediction gains at initially distant conditions, these findings support the framework's ability to acquire information useful across the broader composition--processing domain.

Effective learning was also obtained from modest initial information. Starting with only 10 randomly selected observations, ROSE-HEA attained a mean $R^2$ of 0.7097 after 100 sequential additions. Learning from limited initial information is particularly valuable when experimental constraints restrict initial characterization. The framework also surpassed the strongest conventional predictor while using 32.4\% fewer training observations. These results support a progressive characterization strategy in which an initially small training set is expanded with observations from candidates selected for their expected predictive value. For experiment-intensive HEA development, this offers a practical basis for allocating limited characterization resources and supporting informed composition and processing decisions.

The present evaluation is retrospective and restricted to a fixed candidate pool. Future work should deploy ROSE-HEA in active-learning campaigns where selected alloy--processing conditions are fabricated and tested and new responses guide subsequent decisions. Transfer to additional alloy families and laboratory-level savings remain to be established prospectively. The current diagnostics assess the framework as a whole, leaving individual component effects for targeted ablation studies. Incorporating richer microstructural information and experiment-specific costs could further strengthen its application to practical alloy development.

\section*{CRediT authorship contribution statement}

\textbf{Ahmed Shoyeb Raihan:}
Conceptualization, Methodology, Software, Validation, Formal analysis, Investigation, Data curation, Visualization, Writing -- original draft. \textbf{Md Habibor Rahman:} Investigation, Domain Knowledge, Writing -- review \& editing. \textbf{Ashish Nimbarte:} Resources, Writing -- review \& editing. \textbf{Imtiaz Ahmed:} Conceptualization, Supervision, Investigation, Resources, Project administration, Writing -- review \& editing.

\section*{Data and code availability}

The processed data and code supporting this study are available from the corresponding author upon reasonable request.

\section*{Acknowledgements}

The authors would like to thank the Department of Industrial and Management Systems Engineering (IMSE), West Virginia University (WVU) for their invaluable support.

\bibliographystyle{elsarticle-num}
\bibliography{references}
\end{document}